\documentclass[10pt, conference, compsocconf]{IEEEtran}

\usepackage{times}
\usepackage{epsfig}
\usepackage{graphicx}
\usepackage{amsmath}
\usepackage{amssymb}
\usepackage{floatrow}
\DeclareMathOperator*{\argmin}{arg\,min}
\newcommand{\norm}[1]{\left\lVert#1\right\rVert}
\newcommand{\ignore}[1]{}
\usepackage[pagebackref=true,breaklinks=true,colorlinks,bookmarks=false]{hyperref}

\begin{document}

\title{Interactive 3D Modeling with a Generative Adversarial Network}

\author{Jerry Liu\\
Princeton University\\
\and
Fisher Yu\\
Princeton University \\
\and
Thomas Funkhouser\\
Princeton University\\
}

\maketitle

\begin{abstract}
 We propose the idea of using a generative adversarial network (GAN) to assist users in designing real-world shapes with a simple interface. Users edit a voxel grid with a Minecraft-like interface. Yet they can execute a SNAP command at any time, which transforms their rough model into a desired shape that is both similar and realistic. They can edit and snap until they are satisfied with the result. The advantage of this approach is to assist novice users to create 3D models characteristic of the training data by only specifying rough edits. Our key contribution is to create a suitable projection operator around a 3D-GAN that maps an arbitrary 3D voxel input to a latent vector in the shape manifold of the generator that is both similar in shape to the input but also realistic. Experiments show our method is promising for computer-assisted interactive modeling.
\end{abstract}

\section{Introduction}

There has been growing demand in recent years for interactive tools
that allow novice users to create new 3D models of their own designs.
Minecraft for example, has sold over 120 million copies, up from 20
million just two years ago.

Yet 3D modeling is difficult for novice users.  Current modeling
systems provide either a simple user interface suitable for novices
(e.g., \cite{igarashi,minecraft}) or the ability to make arbitrary
3D models with the details and complexity of real-world objects (e.g.,
\cite{maya,3dsmax}).  Achieving both is an open and fundamental
research challenge.

\begin{figure}[t]
\centering
\includegraphics[width=\columnwidth]{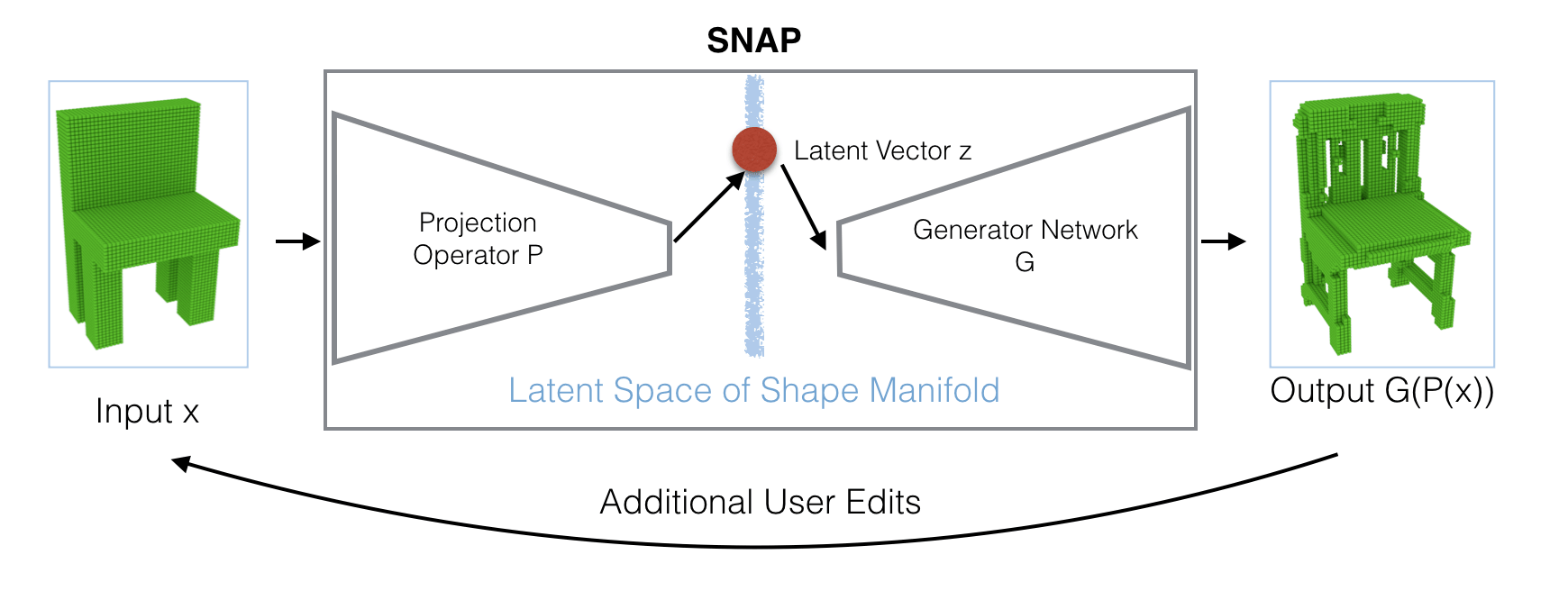}
\vspace*{-5mm}
\caption{Interactive 3D modeling with a GAN. The user iteratively makes edits to a voxel grid with a simple painting interface and then hits a SNAP command to refine the current shape.  The SNAP command projects the current shape into a latent vector shape manifold learned with a GAN, and then generates a new shape with the generator network.  SNAP aims to increase the realism of the user's input, while maintaining similarity.}
\label{fig:workflow}
\end{figure}

In this paper, we investigate how to use Generative Adversarial
Networks (GANs) \cite{goodfellow} 
to help novices create realistic 3D models of their
own designs using a simple interactive modeling tool. 3D GANs have recently been
proposed for generating distributions of 3D voxel grids representing a
class of objects \cite{wu_3dgan}.  Given a latent vector (e.g., a 200-dimensional vector with random values), a 3D-GAN can produce a sample from a latent distribution of voxel grids learned from examples (see the right side of Figure \ref{fig:workflow}). Previous work has used 3D GANs for object classification, shape interpolation, and generating random shapes \cite{wu_3dgan}.  However, they have never before been used for interactive 3D modeling; nor has any other generative deep network. An important limitation with GANs in general has been that while certain subspaces on the manifold generate realistic outputs, there are inherently in-between spaces that contain unrealistic outputs (discussed in Section \ref{sec:approach}). 

\begin{figure*}
\centering
\floatbox[{\capbeside\thisfloatsetup{capbesideposition={right,top},capbesidewidth=4cm}}]{figure}[\FBwidth]
{\caption{A typical editing sequence. The user alternates between painting voxels (dotted arrows) and executing SNAP commands (solid arrows).  For each SNAP, the system projects the current shape into a shape manifold learned with a GAN (depicted in blue) and synthesizes a new shape with a generator network.}
\label{fig:intro-seq}}
{\includegraphics[width=0.7\textwidth]{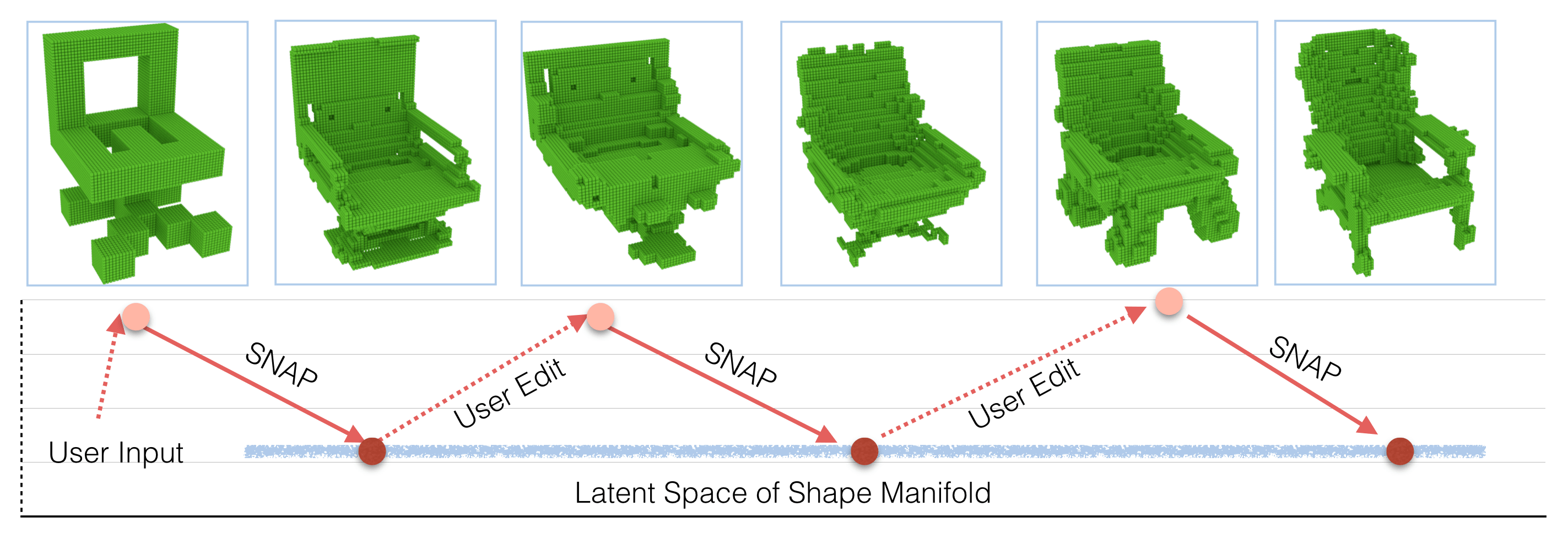}}
\end{figure*}

We propose a model framework around a 3D-GAN which helps hide its weaknesses and allow novice users to easily perform interactive modeling, constraining the output to feasible and realistic shapes. The user iteratively
paints voxels with a simple interface similar to Minecraft \cite{minecraft} and then hits the ``SNAP'' button, which replaces the current voxel grid with a similar one generated by a 3D GAN. 

Our approach is fueled by insights about the disjoint subspaces on the GAN manifold that contain realistic outputs. While there have been various approaches toward a projecting an input into the latent space of a GAN \cite{larsen,zhu}, ours is the first to ensure that the generated output is similar in shape to the input but constrained to the ``good'' spaces of the manifold. This ensures that users are able to generate realistic looking inputs using our GAN framework. The main challenge in implementing such a system is designing this projection operator $P(x)$ from a user-provided 3D voxel grid $x$ to a feature vector $z$ in the latent space of a 3D-GAN (Figure \ref{fig:workflow}).  With such an operator, each SNAP operator can map $x$ to $x'=G(P(x))$, ideally producing an output $x'$
that is not only similar to the input but also representative of real-world
objects in a given training set. We integrate this operator into an interactive modeling tool and demonstrate the
effectiveness of the resulting SNAP command in several typical novice editing sessions.

Figure \ref{fig:intro-seq} depicts an example workflow of this proposed approach. At the beginning, the user sketches the rough shape of an office chair (leftmost panel).  When he/she hits the SNAP button, the system fills in the details of a similar chair generated with a 3D GAN (second panel).  Then the user removes voxels corresponding to the top half of the back, which snaps to a new chair with a lower-back, and then the user truncates the legs of the school chair, which then snaps to a lounge chair with a low base (note that the back becomes reclined to accommodate the short legs). In each case, the user provides approximate inputs with a simple interface, and the system generates a
new shape sampled from a continuous distribution.


The contributions of the paper are four-fold.  First, it is the first
to utilize a GAN in an interactive 3D model editing tool.  Second, it
proposes a novel way to project an arbitrary input into the latent
space of a GAN, balancing both similarity to the input shape and
realism of the output shape.  Third, it provides a dataset of 3D
polygonal models comprised of 101 object classes each with at least $120$ examples in each class, which is the largest, consistently-oriented 3D dataset to date.  Finally, it provides a simple interactive modeling tool for
novice users.


\section{Related Work}

There has been a rich history of previous works on using collections of shapes to assist interactive 3D modeling and generating 3D shapes from learned distributions.  

\vspace*{1mm}\noindent{\bf Interactive 3D Modeling for Novices:}
Most interactive modeling tools are designed for experts (e.g., Maya \cite{maya}) and are too difficult to use for casual, novice users.   To address this issue, several researchers have proposed simpler interaction techniques for specifying 3D shapes, including ones based on sketching curves \cite{igarashi}, making gestures \cite{Zeleznik96}, or sculpting volumes \cite{Galyean91}.   However, these interfaces are limited to creating simple objects, since every shape feature of the output must be specified explicitly by the user.

\vspace*{1mm}\noindent{\bf 3D Synthesis Guided by Analysis:}
To address this issue, researchers have studied ways to utilize analysis of 3D structures to assist interactive modeling.   In early work, \cite{Gal09} proposed an "analyze-and-edit" to shape manipulation, where detected structures captured by wires are used to specify and constrain output models.  More recent work has utilized analysis of part-based templates \cite{cashman,kraevoy}, stability \cite{Averkiou14}, functionality \cite{Nobuyuki15}, ergonomics \cite{Zheng16}, and other analyses to guide interactive manipulation.  Most recently, Yumer et al. \cite{yumer} used a CNN trained on un-deformed/deformed shape pairs to synthesize a voxel flow for shape deformation.  However, each of these previous works is targeted to a  specific type of analysis, a specific type of edit, and/or considers only one aspect of the design problem.  We aim to generalize this approach by using a learned shape space to guide editing operations.

\vspace*{1mm}\noindent{\bf Learned 3D Shape Spaces:}  Early work on learning shape spaces for geometric modeling focused on smooth deformations between surfaces.  For example, \cite{Kilian07}, \cite{Allen03}, and others describe methods for interpolation between surfaces with consistent parameterizations.   More recently, probabilistic models of part hierarchies \cite{Kalogerakis12,huang15} and grammars of shape features \cite{Dang15} have been learned from collections and used to assist synthesis of new shapes.  However, these methods rely on specific hand-selected models and thus are not general to all types of shapes.

\vspace*{1mm}\noindent{\bf Learned Generative 3D Models:}
More recently, researchers have begun to learn 3D shape spaces for generative models of object classes using variational autoencoders \cite{brock,girdhar,sharma} and Generative Adversarial Networks \cite{wu_3dgan}.  Generative models have been tried for sampling shapes from a distribution \cite{girdhar,wu_3dgan}, shape completion \cite{wu_shapenets}, shape interpolation \cite{brock,girdhar,wu_3dgan}, classification \cite{brock,wu_3dgan}, 2D-to-3D mapping \cite{girdhar,rezende,wu_3dgan}, and deformations  \cite{yumer}.  3D GANs in particular produce remarkable results in which shapes generated from random low-dimensional vectors demonstrate all the key structural elements of the learned semantic class \cite{wu_3dgan}. These models are an exciting new development, but are unsuitable for interactive shape editing since they can only synthesize a shape from a latent vector, not from an existing shape.  We address that issue.

\ignore{
\begin{figure}[t!]
\centering
\includegraphics[width=0.4\textwidth]{3dgan_comp.png}
\caption{Taken from 3D-GAN NIPS 2016 slides 
Comparison of results between 3D-GAN and 3D ShapeNets.}
\label{fig:3dgan_comp}
\end{figure}
}

\vspace*{1mm}\noindent{\bf GAN-based Editing of Images}
In the work most closely related to ours, but in the image domain, \cite{zhu} proposed using GANs to constrain image editing operations to move along a learned image manifold of natural-looking images.  Specifically, they proposed a three-step process where 1) an image is projected into the latent image manifold of a learned generator, 2) the latent vector is optimized to match to user-specified image constraints, and 3) the differences between the original and optimized images produced by the generator are transferred to the original image.  This approach provides the inspiration for our project.  Yet, their method is not best for editing in 3D due to the discontinuous structure of 3D shape spaces (e.g., a stool has either three legs or four, but never in between).  We suggest an alternative approach that projects arbitrary edits into the learned manifold (rather than optimizing along gradients in the learned manifold), which better supports discontinuous edits.

\begin{figure}[t]
\centering
\includegraphics[width=1.0\columnwidth]{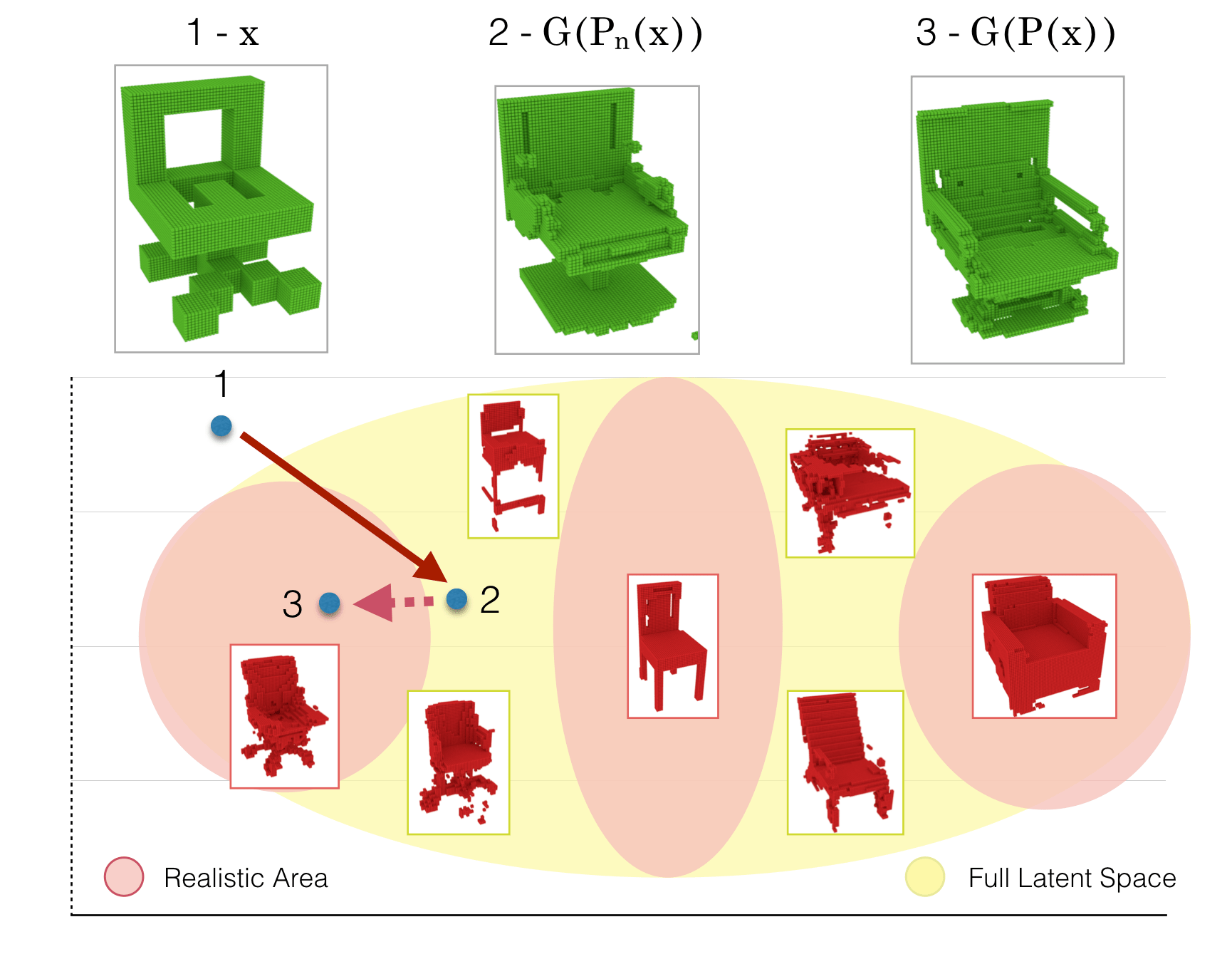}
\caption{Depiction of how subcategories separate into realistic regions within the latent shape space of a generator. Note that the regions in between these modalities represent unrealistic outputs (an object that is in-between an upright and a swivel chair does not look like a realistic chair). Our projection operator $z=P(x)$ is designed to avoid those regions, as shown by the arrows.}
\label{fig:shapespace}
\end{figure}


\section{Approach} \label{sec:approach}

In this paper, we investigate the idea of using a GAN to assist interactive modeling of 3D shapes.  

During an off-line preprocess, our system learns a model for a collection of shapes within a broad object category represented by voxel grids (we have experimented so far with chairs, tables, and airplanes). The result of the training process is three deep networks, one driving the mapping from a 3D voxel grid to a point within the latent space of the shape manifold (the projection operator $P$), another mapping from this latent point to the corresponding 3D voxel grid on the shape manifold (the generator network $G$), and a third for estimating how real a generated shape is (the discriminator network $D$).  

Then, during an interactive modeling session, a person uses a simple voxel editor to sketch/edit shapes in a voxel grid (by simply turning on/off voxels), hitting the ``SNAP'' button at any time to project the input to a generated output point on the shape manifold (Figure \ref{fig:intro-seq}).  Each time the SNAP button is hit, the current voxel grid $x_t$ is projected to $z_{t+1} = P(x_t)$ in the latent space, and a new voxel grid $x_{t+1}$ is generated with $x_{t+1}=G(z_{t+1})$. The user can then continue to edit and snap the shape as necessary until he/she achieves the desired output. 

\begin{figure}[t]
\centering
\includegraphics[width=1.0\columnwidth]{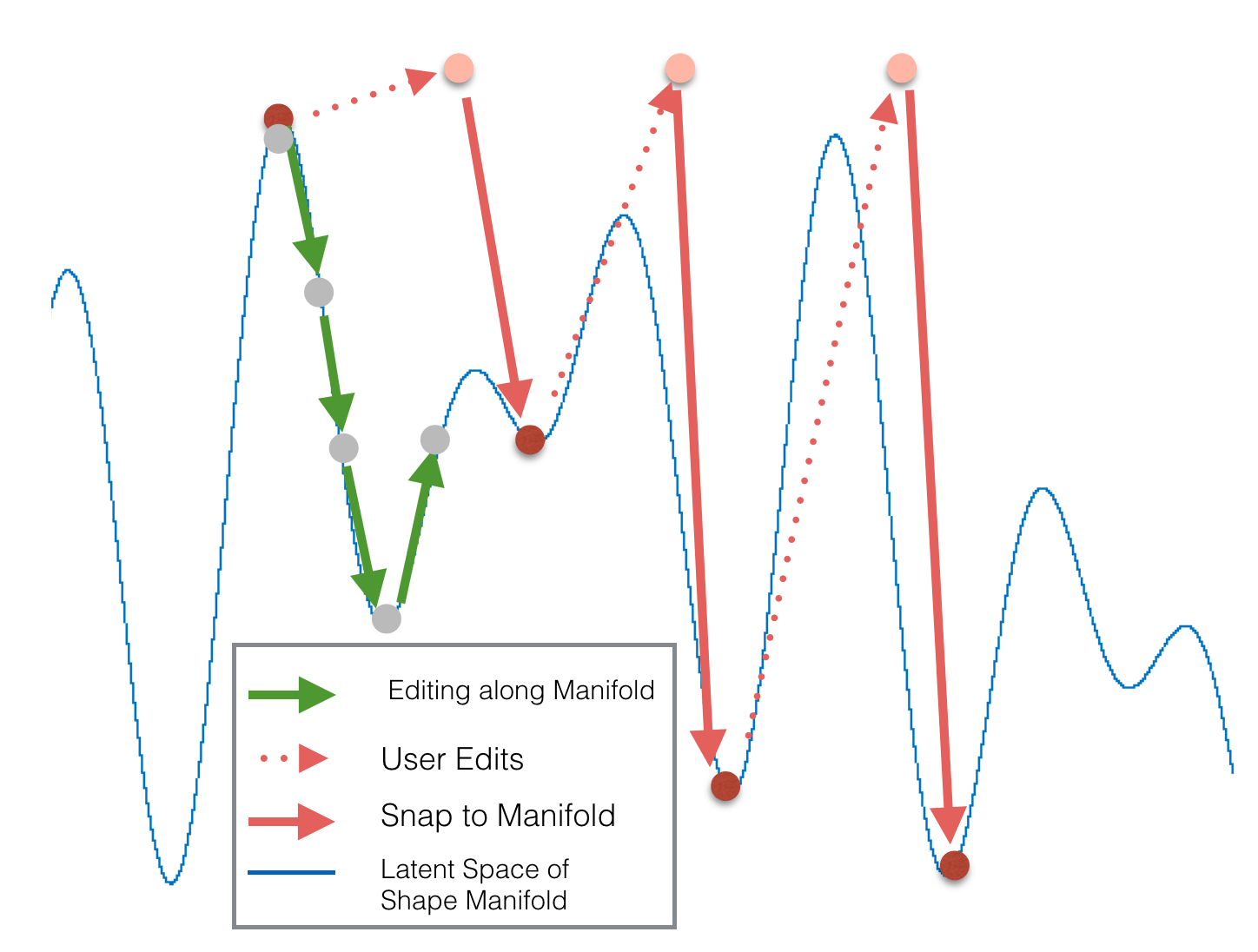}
\caption{Depiction of how the SNAP operators (solid red arrows) project edits made by a user (dotted red arrows) back onto the latent shape manifold (blue curve).  In contrast, a gradient descent approach moves along the latent manifold to a local minimum (solid green arrows).}
\label{fig:offmanifold}
\end{figure}

The advantage of this approach is that users do not have to concern themselves with the tedious editing operations required to make a shape realistic.  Instead, they can perform coarse edits and then ask the system to ``make the shape more realistic'' automatically.   

In contrast to previous work on generative modeling, this approach is unique in that it projects shapes to the ``realistic'' part of the shape manifold {\em after edits are made}, rather than forcing edits to follow gradients in the shape manifold \cite{zhu}.  The difference is subtle, but very significant.  Since many object categories contain distinct subcategories (e.g., office chairs, dining chairs, reclining chairs, etc.), there are modes within the shape manifold (red areas Figure \ref{fig:shapespace}), and latent vectors in the regions between them generate unrealistic objects (e.g., what is half-way between an office chair and a dining chair?).  Therefore, following gradients in the shape manifold will almost certainly get stuck in a local minima within an unrealistic region between modes of the shape manifold (green arrows in Figure \ref{fig:offmanifold}).  In contrast, our method allows users to make edits off the shape manifold before projecting back onto the realistic parts of the shape manifold (red arrows in Figure \ref{fig:offmanifold}), in effect jumping over the unrealistic regions.   This is critical for interactive 3D modeling, where large, discrete edits are common (e.g., adding/removing parts).

\section{Methods}

This section describes each step of our process in detail.  It starts by describing
the GAN architecture used to train the generator and discriminator networks.  It then 
describes training of the projection and classification networks.   Finally, it describes implementation details of the interactive system. 


\subsection{Training the Generative Model} 
\label{sec:generative}

Our first preprocessing step is to train a generative model for 3D shape synthesis.  We adapt the 3D-GAN model from \cite{wu_3dgan}, which consists of a generator $G$ and discriminator $D$. $G$ maps a 200-dimensional latent vector $z$ to a $64 \times 64 \times 64$ cube, while $D$ maps a given $64 \times 64 \times 64$ voxel grid to a binary output indicating real or fake (Figure 5).

\begin{figure}[t]
\centering
\includegraphics[width=0.85\columnwidth]{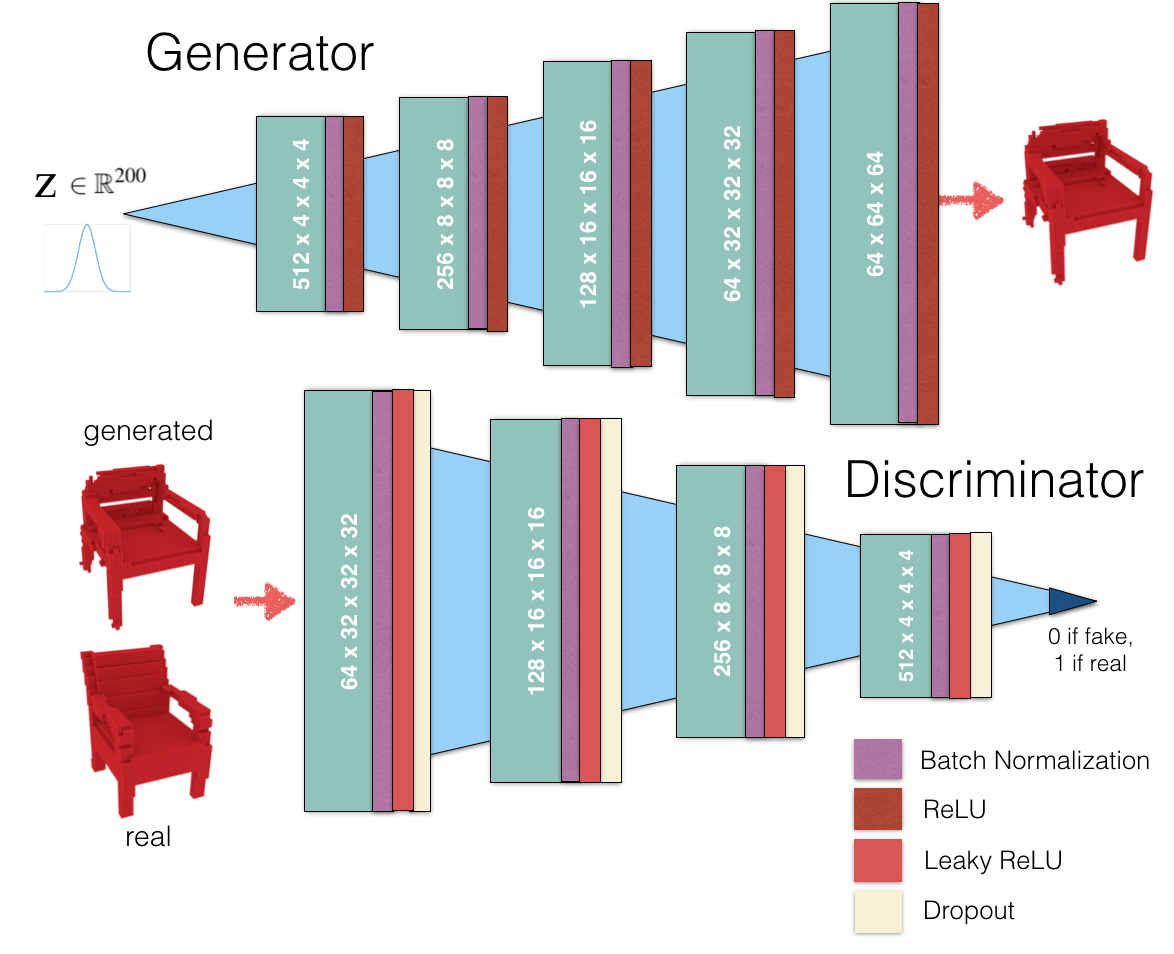}
\vspace*{-2mm}
\caption{Diagram of our 3D-GAN architecture.}
\label{fig:3dgan}
\end{figure}

We initially attempted to replicate \cite{wu_3dgan} exactly, including maintaining the network structure, hyperparameters, and training process. However, we had to make adjustments to the structure and training process to maintain training stability and replicate the quality of the results in the paper. This includes making the generator maximize $\log D(G(z))$ rather than minimizing $\log(1 - D(G(z)))$, adding volumetric dropout layers of $50\%$ after every LeakyReLU layer, and training the generator by sampling from a normal distribution $\mathcal{N}(0,I_{200})$ instead of a uniform distribution $[0,1]$. We found that these adjustments helped to prevent generator collapse during training and increase the number of modalities in the learned distribution. 

We maintained the same hyperparameters, setting the learning rate of $G$ to $0.0025$, $D$ to $10^{-5}$, using a batch size of 100, and an Adam optimizer with $\beta=0.5$. We initialize the convolutional layers using the method suggested by He et al. \cite{he} for layers with ReLU activations.


\subsection{Training the Projection Model} 
\label{sec:projection}

Our second step is to train a projection model $P(x)$ that produces a vector $z$ within the latent space of our generator for a given input shape $x$.   The implementation of this step is the trickiest and most novel of our system because it has to balance the following two considerations:

\begin{itemize}

\item The shape $G(z)$ generated from $z=P(x)$ should be ``similar'' to $x$.  This consideration favors coherent edits matching the user input (e.g., if the user draws rough armrests on a chair, we would expect the output to be a similar chair with armrests). 

\item The shape $G(z)$ must be ``realistic.''  This consideration favors generating new outputs $x'=G(P(x))$ that are indistinguishable from examples in the GAN training set.

\end{itemize}

We balance these competing goals by optimizing an objective function with two terms:
\begin{eqnarray*}
P(x) = \argmin_{z}{E(x, G(z))} \nonumber \\
E(x,x') = \lambda_1 D(x,x') - \lambda_2 R(x') \nonumber
\end{eqnarray*}
where $D(x_1, x_2)$ represents the ``dissimilarity'' between any two 3D objects $x_1$ and $x_2$, and $R(x)$ represents the "realism'' of any given 3D object $x$ (both are defined later in this section).

Conceptually, we can optimize the entire approximation objective $E$ with its two components $D$ and $R$ at once. However, it is difficult to fine-tune $\lambda_1,\lambda_2$ to achieve robust convergence.   In practice, it is easier to first optimize $D(x,x')$ to first get an initial approximation to the input, $z'_0=P_S(x)$, and then use the result as an initialization to then optimize $\lambda_1 D(x,G(z')) - \lambda_2 R(G(z'))$ for a limited number of steps, ensuring that the final output is within the local neighborhood of the initial shape approximation. We can view the first step as optimizing for shape similarity and the second step as constrained optimization for realism. With this process, we can ensure that $G(P(x))$ is realistic but does not deviate too far from the input. 
\begin{eqnarray*}
P_S(x) \leftarrow \argmin_{z}{D(x, G(z))} \nonumber \\
P_R(z) \leftarrow \argmin_{z' | z'_0 = P_S(x)}{\lambda_1 D(x,G(z')) - \lambda_2 R(G(z'))} \nonumber
\end{eqnarray*}

To solve the first objective, we train a feedforward projection network $P_n(x, \theta_p)$ that predicts $z$ from $x$, so $P_S(x) \leftarrow P_n(x, \theta_p)$. We allow $P_n$ to learn its own projection function based on the training data. Since $P_n$ maps any input object $x$ to a latent vector $z$, the learning objective then becomes
$$\sum_{x_i \in X}\min_{\theta_p}{D(x_i, G(P_n(x_i, \theta_p)))}$$
where $X$ represents the input dataset. The summation term here is due to the fact that we are using the same network $P_n$ for all inputs in the training set as opposed to solving a separate optimization problem per input. 

To solve the second objective, \newline $P_R(z) \leftarrow \argmin_{z'}{\lambda_1 D(x,G(z')) - \lambda_2 R(G(z'))}$, we first initialize $z'_0 = P_S(x)$ (the point predicted from our projection network). We then optimize this step using gradient descent; in contrast to training $P_n$ in the first step, we are fine with finding a local minima of this objective so that we optimize for realism within a local neighborhood of the predicted shape approximation. The addition of $D(x,G(z'))$ to the objective adds this guarantee by penalizing the output shape if it is too dissimilar to the input. 

\begin{figure}[t]
\centering
\includegraphics[width=0.9\columnwidth]{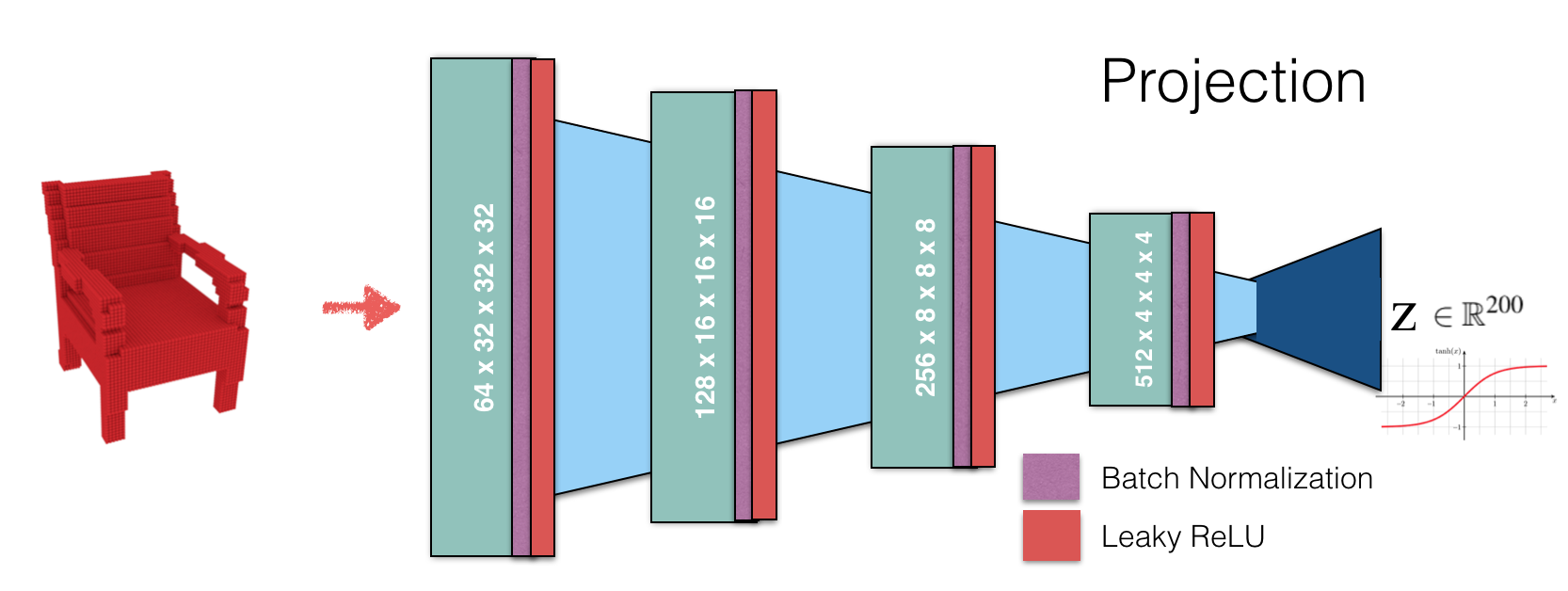}
\caption{Diagram of our projection network. It takes in an arbitrary 3D voxel grid as input and outputs the latent prediction in the generator manifold.}
\label{fig:projection}
\end{figure}

\vspace*{1mm}\noindent{\bf Network Architecture:} The architecture of $P_n$ is given in Figure \ref{fig:projection}. It is mostly the same as that of the discriminator with a few differences: There are no dropout layers in $P_n$, and the last convolution layer outputs a $200$-dimensional vector through a tanh activation as opposed to a binary output. One limitation with this approach is that $z \sim \mathcal{N}(0,1)$, but since $P_n(x) \sim [-1, 1]^{200}$, the projection only learns a subspace of the generated manifold. We considered other approaches, such as removing the activation function entirely, but the quality of the projected results suffered; in practice, the subspace captures a significant portion of the generated manifold and is sufficient for most purposes. 

During the training process, an input object $x$ is forwarded through $P_n$ to output $z$, which is then forwarded through $G$ to output $x'$, and finally we apply $D(x, x')$ to measure the distance loss between $x$ and $x'$. We only update the parameters in $P$, so the training process appears similar to training an autoencoder framework with a custom reconstruction objective where the decoder parameters are fixed. We did try training an end-to-end VAE-GAN architecture, as in Larsen et al. \cite{larsen}, but we were not able to tune the hyperparameters necessary to achieve better results than the ones trained with our method. 

\vspace*{1mm}\noindent{\bf Dissimilarity Function:} The dissimilarity function $D(x_1, x_2) \in \mathcal{R}$ is a differentiable metric representing the semantic difference between $x_1$ and $x_2$.  It is well-known that L2 distance between two voxel grids is a poor measure of semantic dissimilarity.  Instead, we explore taking the intermediate activations from a 3D classifier network \cite{qi,su,maturana,brock}, as well as those from the discriminator. We found that the discriminator activations did the best job in capturing the important details of any category of objects, since they are specifically trained to distinguish between real and fake objects within a given category. We specifically select the output of the $256 \times 8 \times 8 \times 8$ layer in the discriminator (along with the Batch Normalization, Leaky ReLU, and Dropout layers on top) as our descriptor space. We denote this feature space as $conv15$ for future reference.  We define $D(x_1, x_2)$  as $\norm{conv15(x_1) - conv15(x_2)}$.

\begin{figure*}[t]
\centering
\includegraphics[width=0.8\linewidth]{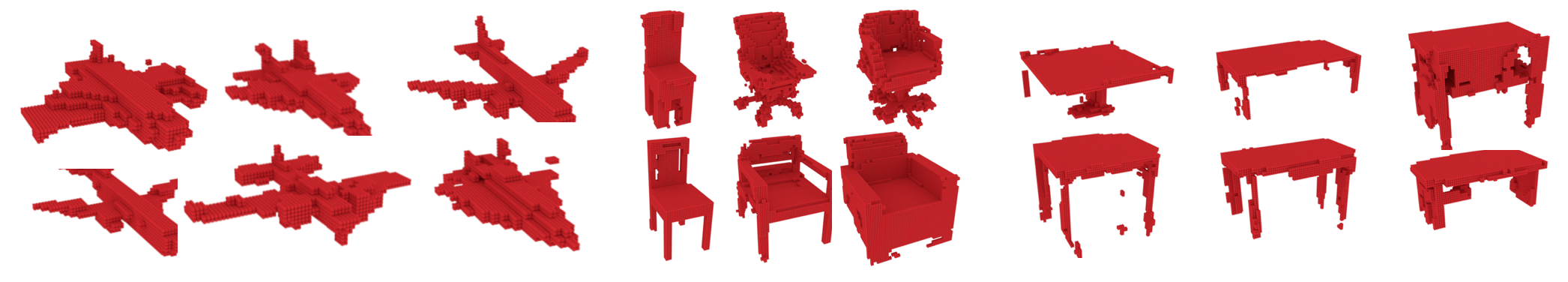}
\vspace*{-4mm}
\caption{Shapes generated from random latent vectors sampled from $\mathcal{N}(0,I_{200})$ using our 3D GANs trained separately on airplanes, chairs, and tables.}.
\label{fig:3dgan-samples}
\end{figure*}

\vspace*{1mm}\noindent{\bf Realism Function:}  The realism function, $R(x) \in \mathcal{R}$, is a differential function that aims to estimate how indistinguishable a voxel grid $x$ is from real object.  There are many options for it, but the discriminator $D(x)$ learned with the GAN is a natural choice, since it is trained specifically for that task.

\vspace*{1mm}\noindent{\bf Training procedure:}  We train the projection network $P_n$ with a learning rate of 0.0005 and a batch size of 50 using the same dataset used to train the generator.   To increase generalization, we randomly drop $50\%$ of the voxels for each input object - we expect that these perturbations allow the projection network to adjust to partial user inputs.


\section{Results}
\label{sec:results}


The goals of these experiments are to test the algorithmic components of the system and to demonstrate that 3D GANs can be useful in an interactive modeling tool for novices. Our hope is to lay groundwork for future experiments on 3D GANs in an 
interactive editing setting.


\subsection{Dataset}   
\label{sec:dataset}

We curated a large dataset of 3D polygonal 
models for this project.   The dataset is largely an extension of the ShapeNet Core55 dataset\cite{chang}, but expanded by 30\% via manual selection of examples from ModelNet40 \cite{wu_shapenets}, SHREC 2014 \cite{li}, Yobi3D \cite{yobi3d} and a private ModelNet repository.  It now covers 101 object categories (rather than 55 in ShapeNet Core55).  The largest categories (chair, table, airplane, car, etc.) have more than 4000 examples, and the smallest have at least 120 examples (rather than 56).  The models are aligned in the same scale and orientation.

We use the chair, airplane, and table categories for experiments in this paper.  Those classes were chosen because they have the largest number of examples and exhibit the most interesting shape variations.


\subsection{Generation Results} 
\label{sec:3dgan-analysis}

We train our modified 3D-GAN on each category separately.  Though quantitative evaluation of the resulting networks is difficult, we study the learned network behavior qualitatively by visualizing results.

\vspace*{1mm}\noindent{\bf Shape Generation:} As a first sanity check, we visualize voxel grids generated by $G(z)$ when $z \in \mathbb{R}^{200}$ is sampled according to a standard multivariate normal distribution for each category.  The results appear in Figure \ref{fig:3dgan-samples}.  They seem to cover the full shape space of each category, roughly matching the results in \cite{wu_3dgan}. 

\begin{figure}[t]
\centering
\includegraphics[width=0.8\columnwidth]{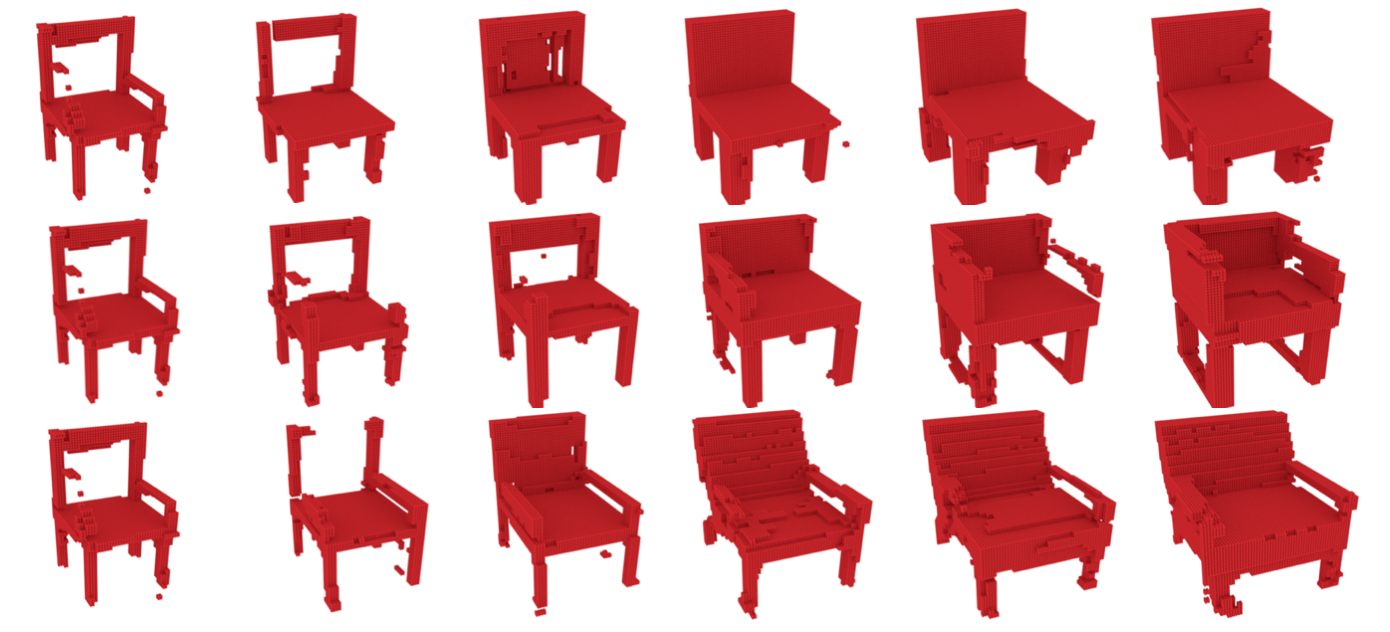}
\caption{Shape interpolation between a randomly sampled reference latent vector $z_r$ and 3 other reference points $z_i$.  The middle images show reconstructions for in-betweens at uniformly spaced interpolations between $z_r$ and $z_i$ in the latent space.}
\label{fig:interp-ring}
\end{figure}


\vspace*{1mm}\noindent{\bf Shape Interpolation:} In our second experiment, we visualize the variation of shapes in the latent space by shape interpolation.
Given a fixed reference latent vector $z_r$, we sample three additional latent vectors $z_0,z_1,z_2 \sim \mathcal{N}(0,I_{200})$ and generate interpolations between $z_r$ and $z_i$ for $0 \leq i \leq 2$. The results are shown in Figure \ref{fig:interp-ring}. The left-most image for row $i$ represents $G(z_r)$, the right-most image represents $G(z_i)$, and each intermediate image represents some $G(\lambda z_r + (1-\lambda) z_i), 0 \leq \lambda \leq 1$. We make a few observations based on these results. The transitions between objects appear largely smooth - there are no sudden jumps between any two objects - and they also appear largely consistent - every intermediate image appears to be some interpolation between the two endpoint images. However, not every point on the manifold appears to be a valid object. For instance, some of the generated chairs are missing legs and other crucial features, or contain artifacts.  This effect is particularly pronounced when $z_r$ and $z_i$ represent shapes with extra/missing parts or in different subcategories.  This result confirms the need for the realism term in our projection operation.

\subsection{Projection Results} 
\label{sec:projection-results}

In our next set of experiments, we investigate how well the projection operator predicts the latent vector for a given input shape.   

\begin{figure}[t]
\centering
\includegraphics[width=\columnwidth]{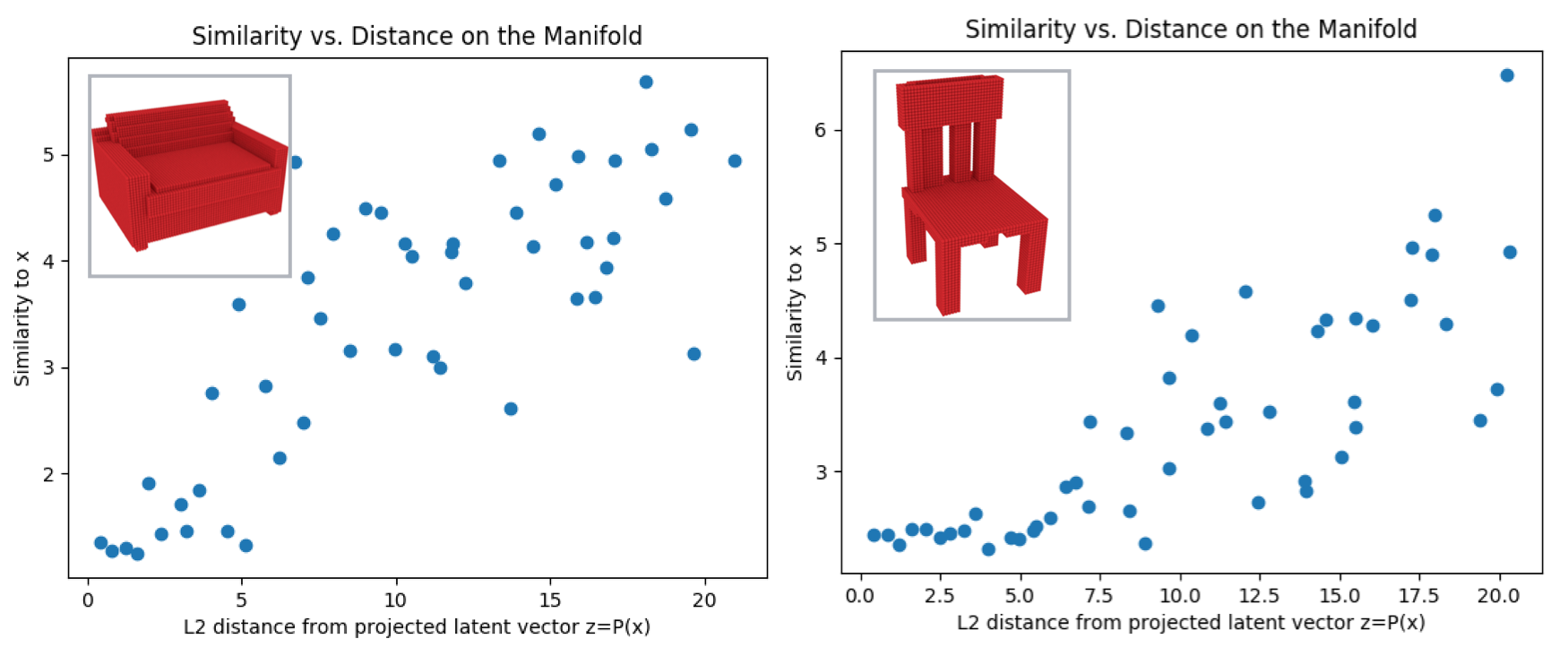}
\caption{Graph showing the correlation between the L2 distance in the latent space and the similarity measure $D(G(z), x)$.  The horizontal axis shows L2 distances from $z=P(x)$, where $x$ is sampled from the training set. The vertical axis shows the similarity measure $D(G(z), x)$.   Note that they are largely correlated.}
\label{fig:sim-dist-proj}
\end{figure}

Each projected vector $P_n(x)$ appears to find an optimum of the distance function within a wide local radius on the latent space with respect to the input $x$. This is demonstrated in Figure \ref{fig:sim-dist-proj}. We measure $D(G(z), x)$ with respect to the distance of $z$ from $P(x)$. We sample various inputs from the training set. We note that $D(G(z), x)$ is still highly non-smooth and non-convex, but the projected point $P(x)$ is able to achieve a rough local minimum. This means that our projection network is adequately finding an approximately optimal point in the $conv15$ feature space given an input. 

\begin{figure*}[t]
\centering
\floatbox[{\capbeside\thisfloatsetup{capbesideposition={left,top},capbesidewidth=3.0cm}}]{figure}[\FBwidth]
{\caption{Examples of chairs projected onto the generated manifold, with their generated counterparts shown as the output. The direct output of the projection network $P_n$is shown in the second row, while the output of the full projection function $P$ is shown in the last row. }
\label{fig:proj-generate-samples}}
{\includegraphics[width=0.8\textwidth]{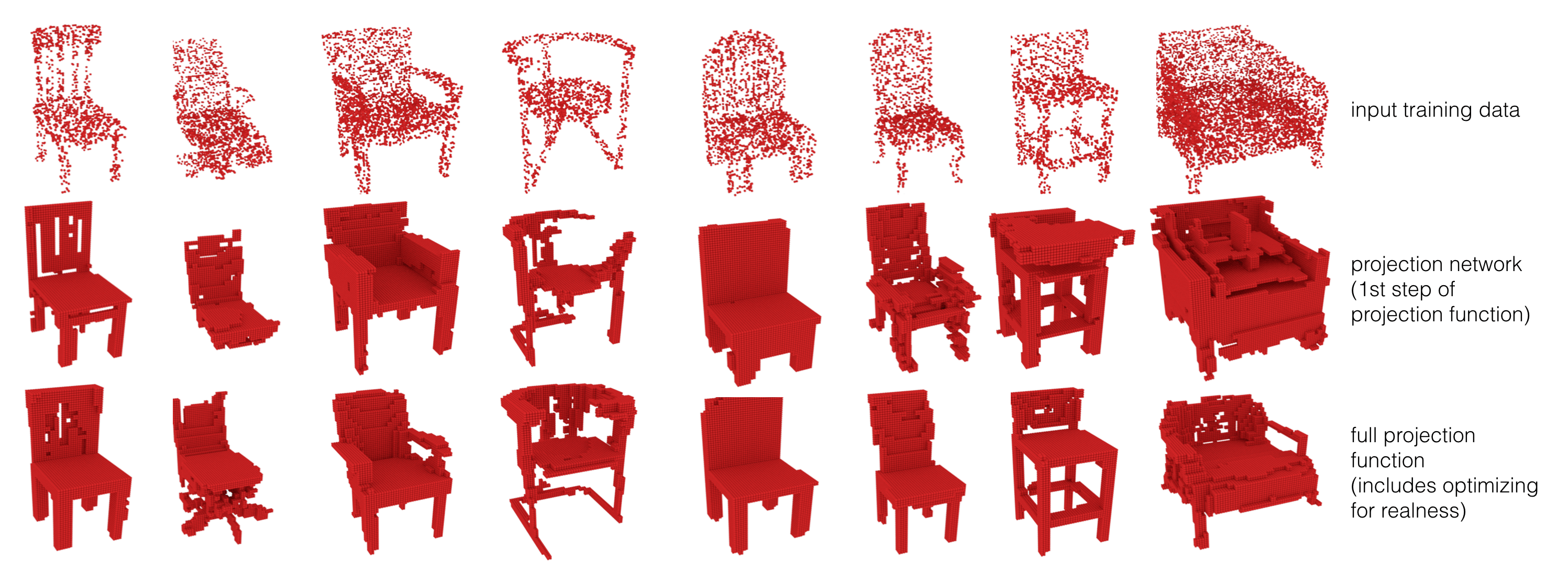}}
\end{figure*}

A direct visual comparison of input and projected samples of chairs is demonstrated in Figure \ref{fig:proj-generate-samples}. An input chair $x$ is provided in the first row (with many voxels missing). The second row shows the generated representation of the predicted latent vector from the projection network $P_n(x)$. The third row adds the second step of the projection function, which optimizes the latent vector towards a point $P(x)$ that would be classified as real by the discriminator. 

On the whole, we see that the direct output of the projection network in the second row, $G(P_n(x))$, maintains the general visual features of the input. For instance, the the height of the back in the first column and the shape of the armrests in the third column are preserved. However, many of the generated images either contain missing components or contain extra artifacts which detract from the visual quality. The chairs in the 3rd and 4th images appear incomplete, while the chairs in the 7th and 8th images appear too noisy.  

The output of the full projection operator shown in the third row address most of these issues.  The second optimization step of the projection operator that pushes the predicted latent vector into a more realistic region of the manifold  creates a noticeable improvement in the visual quality of the results overall.  For example, in the second column, the final swivel chair looks more realistic and better match the style of the input than the fragmented prediction of the projection network alone. Of course, there are cases where coercing realism moves the shape away from the user's intended edit (e.g., the couch in the last column is transformed into a wide chair with armrests).  The trade-off between realism and faithfulness to the user's edit could be controlled with a slider presented to the user to address this issue.

\begin{figure}[t]
\centering
\includegraphics[width=\columnwidth]{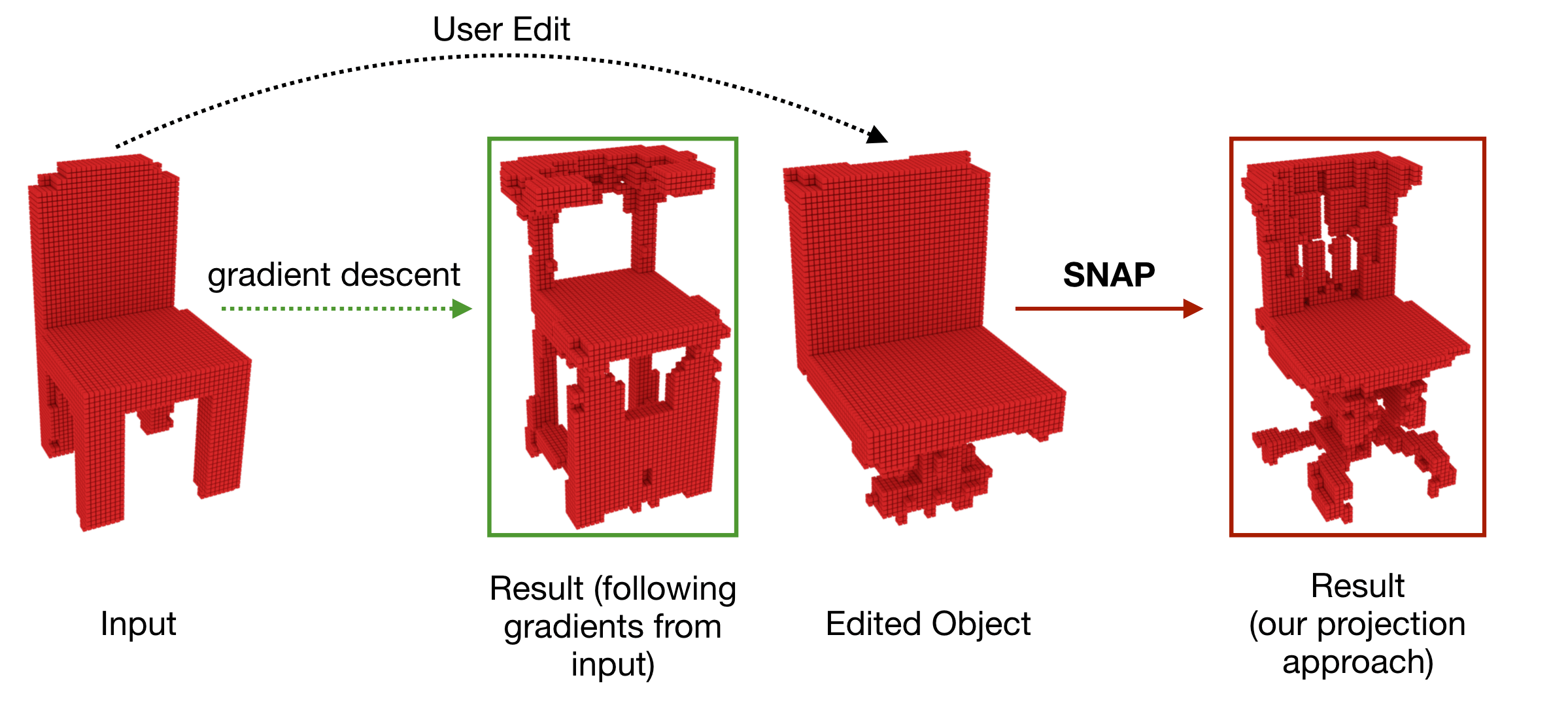}
\caption{Comparison of our projection function with the gradient approach, as discussed in Section \ref{sec:approach}.}
\label{fig:comp-projection-func}
\end{figure}

Fig. \ref{fig:comp-projection-func} highlights the advantages of our hybrid projection approach compared to a pure gradient approach, as mentioned in Section \ref{sec:approach}. As seen, the gradient approach converges in an unrealistic region of the manifold, creating an unrealistic chair. In the meantime, our approach directly projects the edited object back into a distinct, realistic region of the manifold - as a result the desired swivel chair appears much more realistic. 

\section{Shape Manipulation Application}
\label{sec:application}

In this section, we describe how the 3D GAN and projection networks are integrated into an interactive
3D shape manipulation application.

\begin{figure}[t]
\centering
\includegraphics[width=0.9\columnwidth]{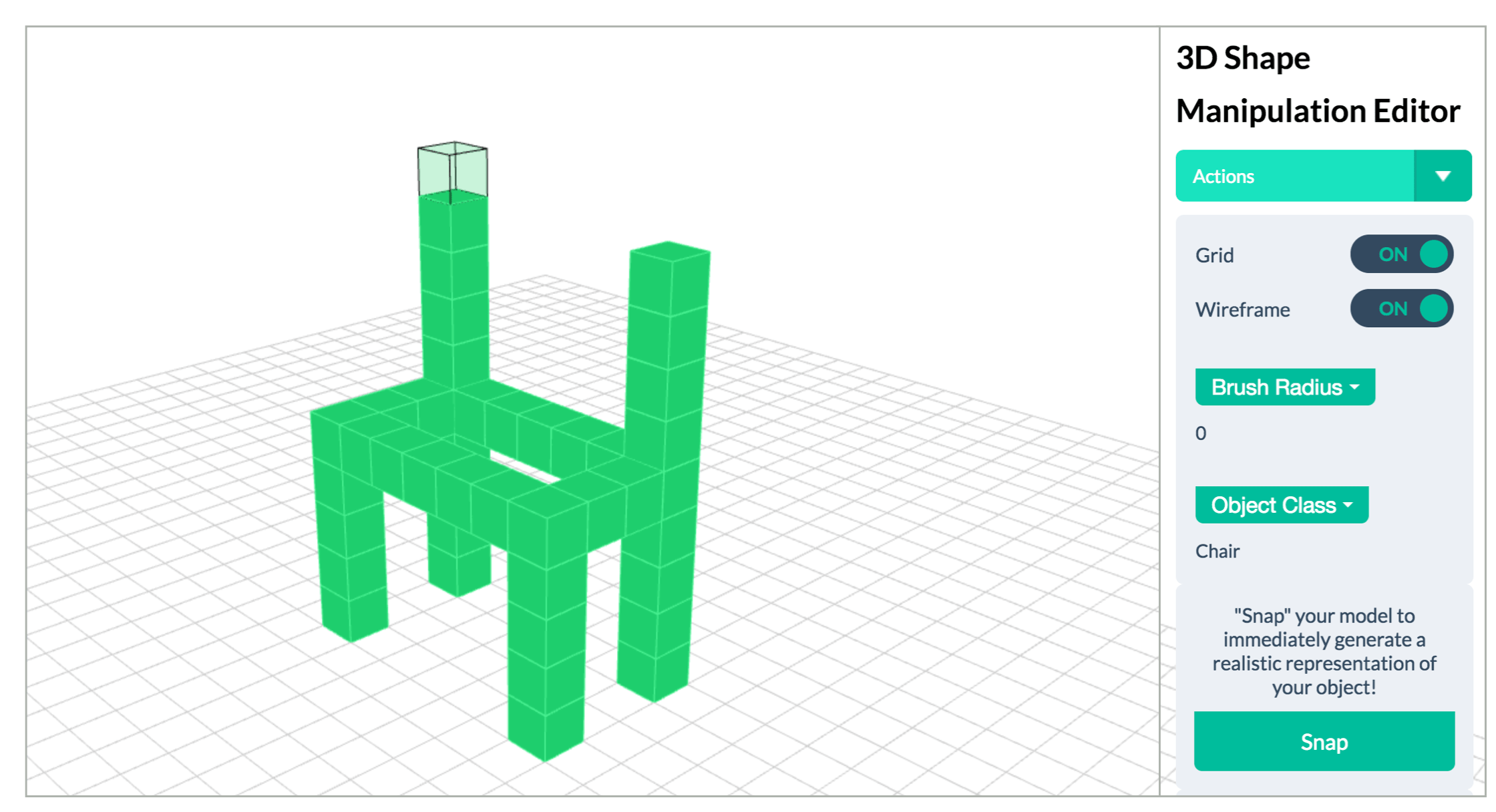}
\caption{Voxel editing interface, adopted from Voxel Builder.}
\label{fig:interface-outline}
\end{figure}

The application is based upon an open-source Voxel Builder tool \cite{voxbuilder}, which provides a user interface for easily creating and editing voxels in a grid (Figure \ref{fig:interface-outline}). We customize the source code by removing the default editing operations and replacing them with a single SNAP button.  When the user hits that button, the current voxel grid is projected into the latent shape manifold and then forwarded through the generator to create a new voxel grid that lies on the manifold. The user iterates between editing voxels and hitting the SNAP button until he/she is happy with the result.

We postprocess the voxels on the server end before returning them to the user. This is an important step to improve the quality and realism of the generated results.  Specifically, we remove small connected components of voxels from the output.  For symmetric objects, we generate only half of the output and then synthesize the other half with a simple reflection.  These simple steps improve both the speed and realism of the generated outputs. 


\ignore{
\begin{figure}[t]
\centering
\includegraphics[width=0.4\textwidth]{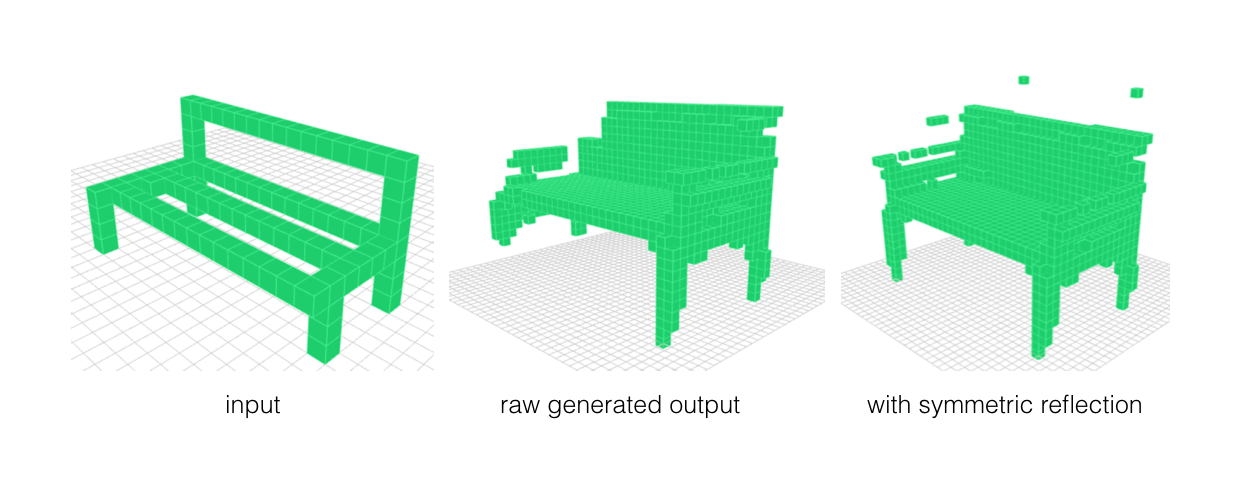}
\caption{Examples of visual quality with and without symmetric reflection.}
\label{fig:symmetry-example}
\end{figure}
}


\begin{figure*}[t]
	\centering
	\includegraphics[width=1.0\textwidth]{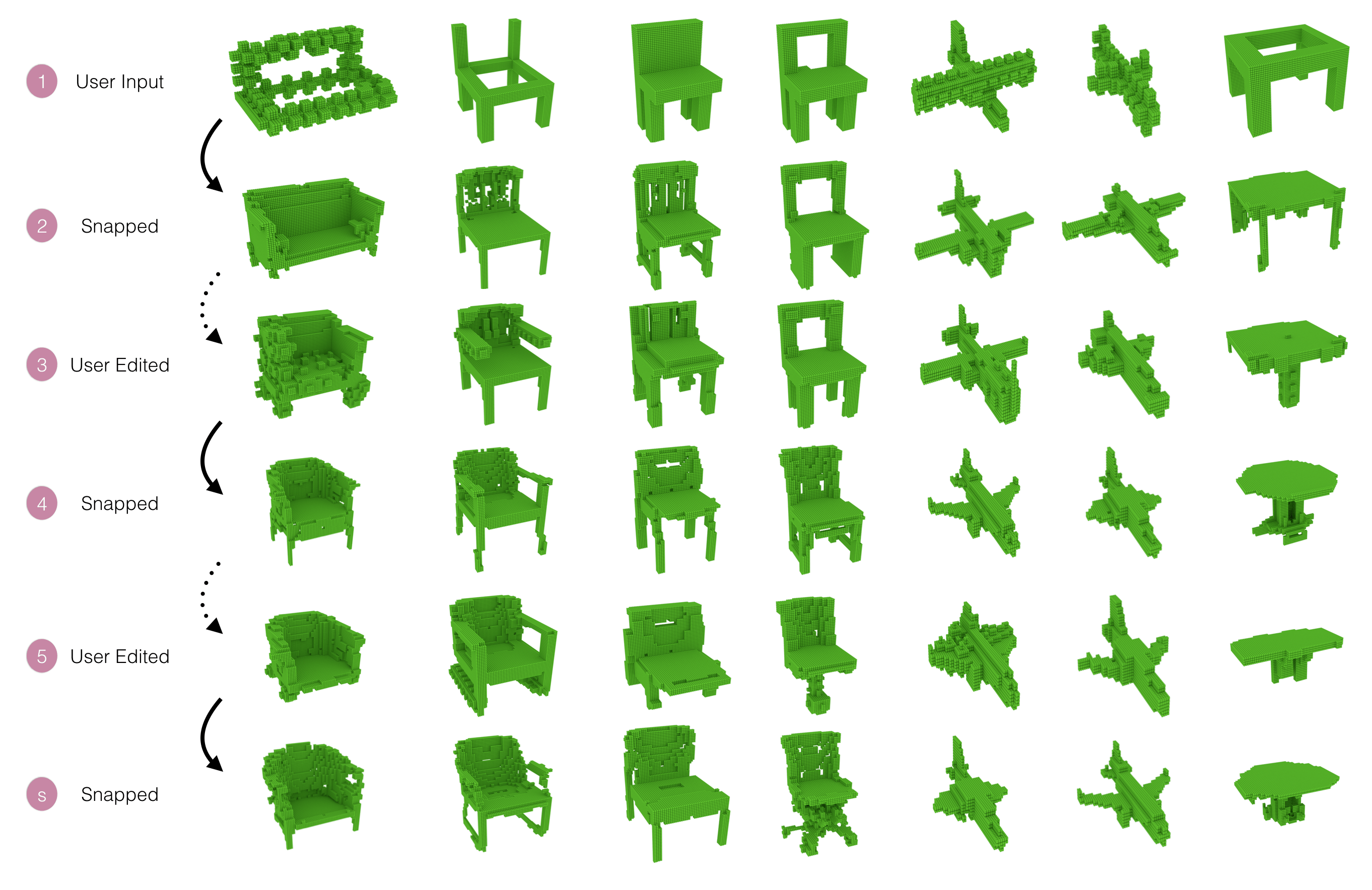}
    \vspace*{-4mm}
	\caption{Demonstration of editing sequences performed with our interface. The user paints an initial shape (top) and then alternates between snapping it (solid arrows) and adding/removing voxels (dotted arrows).  After each snap, the resulting object conforms roughly to the specifications of the user. }\label{fig:interface-continued}
\end{figure*}

The speed of a complete SNAP operation is around 9 seconds on average using an NVIDIA Tesla M40 GPU for the forward passes of the networks and gradient optimization steps. This may be a bit too slow for a production-level interactive modeling tool; however, our goal is to investigate the idea of using a GAN for 3D modeling, not to provide a deployable system for commercial users. 

\vspace*{1mm}\noindent{\bf Editing Sequence Results:}
Our final set of experiments show the types of edits that are possible with the assistance of a 3D GAN.  In each experiment, we show a sequence of voxel edits followed by SNAP commands and then analyze whether/how the SNAP assists the user in creating detailed and realistic models of their own design.



Figure \ref{fig:interface-continued} shows several editing sequences comprising multiple voxel edits and SNAP commands.   Results are shown for chairs, airplanes, and tables. For each editing sequence, the user starts by creating/snapping an object from scratch (top row), and then continues to edit the generated object by adding or removing voxels (dotted arrows) and then snapping (solid arrows) for three iterations.  We can see that the snapped objects are generally similar to their input, but more realistic representations of the object class.  For example, it fills in the details of the couch in the first snap in the first column, and fixes the aspect ratio of the chair in the last snap of the third column.

The snap operator often adjusts the overall style of the object to accommodate user edits.  For example, in the first column, the user shrinks the width of the couch, and the snapped result is no longer rectangular - it becomes a fauteuil-esque chair with high armrests and a curved back. Shortening the wings of a plane in the sixth column causes the overall frame to transform into a sleek fighter jet. 
This implies that our approach is able to find a good balance between similarity and realism, returning results for the user that match both the edits made by the user as well as the style of a realistic object. 





\begin{figure}[t]
\centering
\includegraphics[width=1.0\columnwidth]{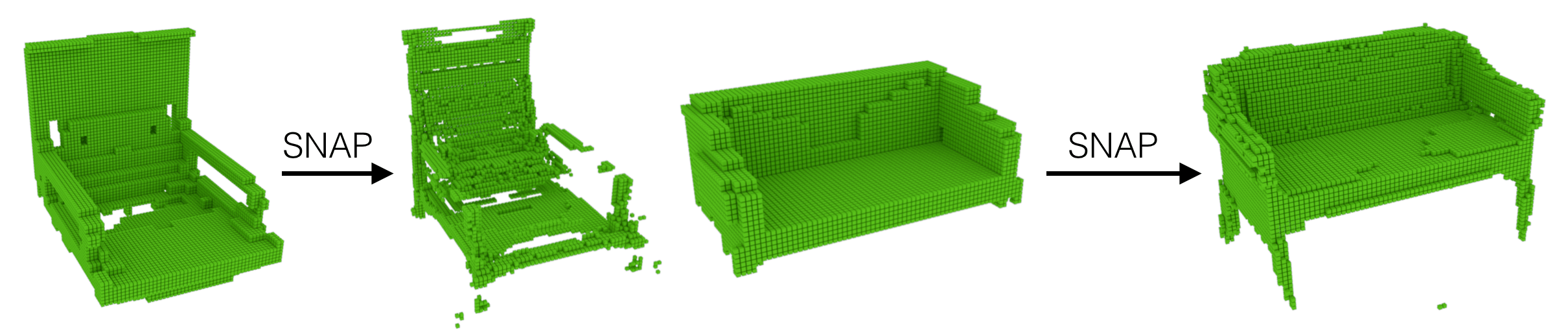}
\caption{Failure cases.  Note the snapped result in the left sequence is unrealistic, and the result in the right sequence adds legs to what is intended as a sofa.}
\label{fig:failure-cases}
\end{figure}

\vspace*{1mm}\noindent{\bf Failure Cases:}
There are some cases where the SNAP operator makes the result worse rather than better.  It might produce results that are unrealistic (left pair in Figure~\ref{fig:failure-cases}), perhaps because the GAN has limited training data.   Or, it might produce results dissimilar from the user intentions (right pair in Figure~\ref{fig:failure-cases}), perhaps because realism is weighted too highly in the projection operator.  These failures could be mitigated somewhat with more sophisticated validation and/or post-processing of generated outputs.  We did not investigate such methods, as they would only mask the conclusions that can be made from our results.

\section{Conclusion}

In summary, we present a novel means of performing 3D shape manipulation by developing a framework of deep learning models around a deep generative model. We use 3D-GAN as our generative model, and design a projection function to project user input to a latent point in the manifold that both captures the input shape and appears realistic. In the process we employ our discriminator to provide a feature space as well as a measurement of realism which is essential towards improving the quality of our results.  We've shown that editing operations with an easy-to-use interface can invoke complex shape manipulations adhering to the underlying distribution of shapes within a category.

This work is just a baby step towards using generative adversarial networks to assist interactive 3D modeling.  We have investigated the core issues in developing a system based on this idea, but it may be years before GAN-based systems produce outputs of quality high enough for production systems. Future work should develop better ways to learn projection and generation operators, and investigate alternative strategies for balancing trade-offs between matching user inputs and prior shape distributions.

\pagebreak
{\small
\bibliographystyle{ieee}
\bibliography{references}
}

\end{document}